\documentclass[11pt]{article}

\usepackage[preprint]{acl}

\usepackage{times}
\usepackage{latexsym}

\usepackage[T1]{fontenc}

\usepackage[utf8]{inputenc}

\usepackage{microtype}

\usepackage{inconsolata}

\usepackage{graphicx}

\usepackage{times}
\usepackage{latexsym}
\usepackage{arydshln}
\usepackage{multirow}
\usepackage[table]{xcolor}
\usepackage{float}
\usepackage{booktabs}
\usepackage{longtable}
\usepackage{caption}
\usepackage{makecell}
\usepackage{xcolor}
\usepackage{amsmath}

\definecolor{nicegreen}{RGB}{40, 164, 40}
\definecolor{nicered}{RGB}{199, 35, 40}
\definecolor{niceblue}{RGB}{22, 48, 194}

\usepackage[T1]{fontenc}
\usepackage[utf8]{inputenc}

\usepackage{stfloats}
\usepackage{hyperref}
\usepackage{microtype}
\usepackage{inconsolata}
\usepackage{graphicx}
\usepackage[inline]{enumitem}

\newcommand{\xhdr}[1]{\noindent {\bf #1.} }

\usepackage{
booktabs,
longtable,
tabularx,
siunitx,
adjustbox,
colortbl,
xcolor,
pdflscape
}

\title{Compression of Emotion in Language Model Representations}

\author{Laya Iyer \\
  Stanford University \\
  \texttt{laya@stanford.edu} \\\And
  Sanmi Koyejo \\
  Stanford University \\
  \texttt{sanmi@stanford.edu} \\\And
  Chen Shani \\
  Tel Aviv University\\
  \texttt{cshani@tau.ac.il}}

\begin{document}
\maketitle
\begin{abstract}

Large language models (LLMs) have been evaluated on their ability to identify emotions expressed in text and produce fluent emotional language. However, output predictions alone do not reveal how fine-grained emotion distinctions are organized internally. We study seven LLMs on GoEmotions, an English Reddit dataset annotated with 27 fine-grained emotion categories plus neutral. We compare model predictions and emotion-prompted final-token representations with the 28 GoEmotions categories and with coarser emotion-family and sentiment groupings.

We find a consistent pattern of \textit{overcompression} relative to the
fine-grained GoEmotions taxonomy. Under the tested KMeans protocol, when we cluster their hidden representations without using emotion labels, the clearest split is into two groups that largely separate positive from negative emotion. Coarser emotion groupings are also recovered with fewer clusters than the full 28-category taxonomy. Yet linear probes retain substantial fine-grained predictive information. These results suggest that fine-grained emotion information remains accessible even though broader affective structure more strongly organizes the unsupervised geometry.
\end{abstract}



\section{Introduction}
Large language models (LLMs) can often identify emotions expressed in text, but successful recognition does not tell us how those distinctions are organized in their internal representations. Human emotion can be described at different levels of granularity: an utterance may be broadly negative, belong to a fear-related family, or express a more specific emotion such as \textit{nervousness}. We ask whether LLM representations preserve these fine-grained distinctions, or whether they organize emotional utterances at broader levels.

For example, when prompted to identify the emotion expressed by the utterance ``Btw--I love your flair!'', whose GoEmotions annotation is \textit{love}, \texttt{Llama-3.2-1B-Instruct} responds ``Positive Emotional State.'' This response could reflect several different failures: the model may recognize the emotion but use a broader label, fail to map its interpretation onto the requested vocabulary, or represent the utterance itself at a broader level. Output behavior alone cannot distinguish among these possibilities.

\begin{figure*}[t]
    \centering
    \includegraphics[width=\textwidth]{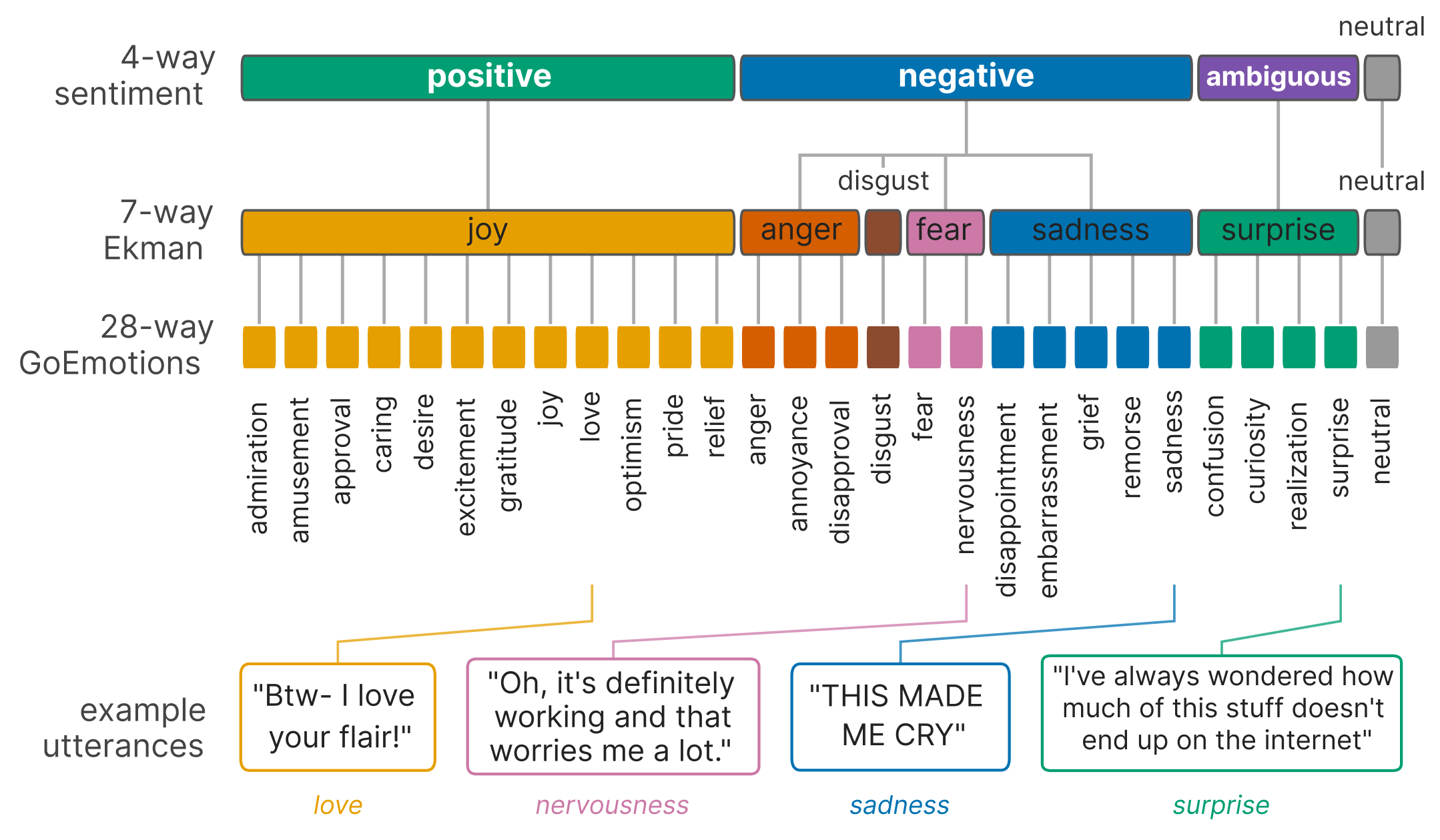}
    \caption{\textbf{Three levels of granularity in the GoEmotions label taxonomy.} The original 27 emotion categories plus neutral are mapped to six
    Ekman-style emotion families plus neutral, and to four sentiment classes: positive, negative, ambiguous, and neutral. The same 36,308 utterances are evaluated under each label grouping, allowing us to compare fine-grained and coarse emotion structure.}
    \label{fig:taxonomy}
\end{figure*}

We therefore study emotion representation from three complementary perspectives. First, \textit{output classification} asks whether a model can explicitly name the annotated emotion. Second, \textit{geometric separation} asks whether utterances with the same human-defined emotion labels form distinct regions in the model's hidden representations without using those labels during clustering. Third, \textit{supervised recoverability} asks whether fine-grained emotion labels can still be predicted from those representations using a linear probe. These measurements need not agree. For example, a label can remain recoverable even when examples from that category do not form a sharply separated cluster. Our central question is therefore: \textbf{How well do fine-grained GoEmotions labels organize task-conditioned hidden-state geometry, and how much fine-grained information remains recoverable when unsupervised separation is weak?}

We evaluate seven open instruction-tuned models from the Llama, Gemma, and Qwen families on 36,308 GoEmotions examples with a single released aggregated label. We compare the original 28 emotion categories with seven broader emotion families and four sentiment classes, and analyze both model outputs and hidden representations across layers. Under the tested KMeans protocol, the strongest global separation is coarse: silhouette is highest at \(k=2\) across all seven models, and these two-cluster partitions are consistently related to positive-versus-negative valence. Coarser taxonomies also reach near-maximal label alignment with far fewer clusters than the 28-way taxonomy, while linear probes retain substantial fine-grained predictive information. We refer to this pattern as \textit{overcompression} relative to the GoEmotions taxonomy. Broader affective structure more strongly organizes the unsupervised geometry even though fine-grained emotion information remains accessible.

\section{Related Work}

Human emotion has been modeled at multiple levels of granularity: as a small set of basic categories \citep{ekman1992,plutchik1980}, as dimensions such as valence and arousal \citep{russell1980}, and as a richer set of continuously related states \citep{cowen2017selfreport}. Psychology describes the ability to distinguish similar emotional states as \emph{emotional granularity} \citep{barrett2001knowing,kashdan2015unpacking}. GoEmotions operationalizes a fine-grained taxonomy in NLP through 27 emotion categories plus neutral \citep{demszky2020goemotions}. The authors arrived at these categories through multiple rounds of annotation, removing emotions that were rare or difficult to distinguish and also making sure the 27 emotion dimensions were significantly dissociable under their validation analysis \citep{demszky2020goemotions}. Importantly, the original analysis also finds coarse structure in human emotion judgments: hierarchical clustering recovers higher-level sentiment groupings, and a BERT baseline performs better on coarser Ekman-style and sentiment groupings than on the full taxonomy \citep{demszky2020goemotions}. Thus, overlap among nearby fine-grained labels is not by itself evidence of a model-specific failure.

Recent work shows that emotion-related information is present in language-model representations. Emotion-relevant representations can causally affect behavior
\citep{tak2025mechanistic}, neuron populations can exhibit emotion selectivity \citep{lee2025emotionneurons}, and low-dimensional affective structure can be recovered from hidden states
\citep{reichman2026emotions,sofroniew2026emotion}.
\citet{reichman2026emotions} derive an emotional subspace using synthetically rewritten sentences and show that it transfers to real-world emotion datasets;
we instead study how naturally occurring Reddit utterances are organized relative to fine-grained labels. \citet{okawa2026hierarchical} find hierarchical organization in model output probabilities and increasingly complex hierarchies in larger models. Our analysis instead studies unsupervised hidden-state geometry and linear recoverability across a smaller, heterogeneous set of model sizes.

Finally, supervised probes and unsupervised geometry answer different questions. Probe performance establishes what information can be recovered
from a representation, but does not by itself establish how that information is organized \citep{hewitt2019probes}. Contextual representations can also be strongly anisotropic, motivating care when interpreting their geometric structure \citep{ethayarajh2019contextual}. We therefore distinguish output classification, unsupervised geometric separation, and supervised recoverability. Prior work establishes that emotion information exists in LLMs; we ask how finely naturally occurring emotional text is organized relative to the GoEmotions taxonomy.

\section{Data \& Models}

We describe the emotion dataset, target taxonomies, and language models used throughout our experiments.

\subsection{Dataset}
We use GoEmotions, a dataset of English Reddit comments annotated with 27 fine-grained emotion categories plus neutral \citep{demszky2020goemotions}. We retain examples assigned exactly one released emotion label in train set, reducing the dataset from 43,410 to 36,308 examples. We apply no additional deduplication. These examples have a single aggregated dataset label.

The resulting dataset is substantially class-imbalanced. \textit{Neutral} is the most common category, with 12,823 examples, while \textit{grief}, \textit{pride}, \textit{nervousness}, and \textit{relief} contain only 39, 51, 85, and 88 examples, respectively. We therefore report macro-averaged output and probe metrics in addition to aggregate accuracy, and stratify the supervised-probe splits by the original 28-way label. Class imbalance can still affect unsupervised clustering, however, because frequent categories contribute more examples to the fitted geometry.

We evaluate the same examples under three label granularities shown in \hyperref[fig:taxonomy]{Figure~\ref{fig:taxonomy}}. The fine-grained taxonomy contains the original 27 GoEmotions categories plus neutral. Following the mappings released with GoEmotions, we group these into six Ekman-style emotion families plus neutral, giving seven categories, and into four sentiment classes: positive, negative, ambiguous, and neutral. Here, \textit{ambiguous} is a sentiment grouping in the dataset rather than a measure of annotation uncertainty. These nested label spaces let us compare how readily fine-grained and broader emotion structure is recovered in model outputs and representations.

\subsection{Models}

We evaluate seven open instruction-tuned language models from three model families, ranging from approximately 1B to 9B parameters:
\begin{itemize}
    \item \texttt{Llama-3.2-1B-Instruct} (16 Transformer blocks) \citep{meta2024llama32}
    \item \texttt{Llama-3.2-3B-Instruct} (28 Transformer blocks) \citep{meta2024llama32}
    \item \texttt{Llama-3.1-8B-Instruct} (32 Transformer blocks) \citep{grattafiori2024llama3}
    \item \texttt{Gemma-2-2B-IT} (26 Transformer blocks) \citep{gemmateam2024gemma2}
    \item \texttt{Gemma-2-9B-IT} (42 Transformer blocks) \citep{gemmateam2024gemma2}
    \item \texttt{Qwen2.5-3B-Instruct} (36 Transformer blocks) \citep{yang2024qwen25}
    \item \texttt{Qwen2.5-7B-Instruct} (28 Transformer blocks) \citep{yang2024qwen25}
\end{itemize}

We focus on open-weight models because our central analyses require access to intermediate hidden states. We evaluate all models on the same examples using the same task prompts and representation-extraction procedure, with each model's native tokenizer and chat template. Because the models differ in architecture, training, and model family, comparisons by parameter count are descriptive rather than a controlled scaling analysis. Layer locations are reported by both absolute layer and relative depth.

For the latent-space experiments, we extract one representation for each example at every hidden-state layer, including the embedding representation as layer 0. Our primary cross-model representation is the hidden state at the final input position after applying the emotion-identification prompt and the model's chat template. We compare this choice against raw-text inputs and alternative pooling strategies in \hyperref[sec:representation-extraction]{Section~\ref{sec:representation-extraction}}; all primary cross-model analyses use the emotion-prompted final-position representation.

\section{Experimental Setting}
We evaluate emotion representation from three complementary perspectives. First, the output-space analysis (\hyperref[sec:output-space]{Section~\ref{sec:output-space}}) measures whether models can explicitly identify the annotated emotion. Second, the latent-space analysis (\hyperref[sec:latent-space]{Section~\ref{sec:latent-space}}) asks whether utterances with the same emotion labels form distinct groups in hidden-state space without using those labels during clustering. Third, linear probes test how much label information remains recoverable even when the categories are not cleanly separated geometrically.

\subsection{Output Space Analysis}
\label{sec:output-space}

We evaluate explicit emotion prediction under four conditions: (1) \textbf{28-way, no list}, where the model predicts an emotion without being shown candidate labels; (2) \textbf{28-way, list given}, where all 28 GoEmotions labels are provided; (3) \textbf{7-way}, using the six Ekman-style emotion families plus neutral; and (4) \textbf{4-way sentiment}, using positive, negative, ambiguous, and neutral. The first two conditions vary the information provided in the prompt while holding the target taxonomy fixed, whereas the latter two use progressively coarser label spaces (\hyperref[fig:taxonomy]{Figure~\ref{fig:taxonomy}}). Full prompts are provided in \hyperref[sec:prompts]{Appendix~\ref{sec:prompts}}.

We parse each generated response into the requested label space using case-normalized exact matching, a small synonym mapping for the no-list condition, and substring matching; responses that cannot be mapped are marked as \emph{unknown}. Unknown predictions are counted as incorrect when computing accuracy and as errors when computing macro F1. We report accuracy, macro F1, and unknown-response rate. Full parsing rules appear in \hyperref[sec:output-parsing]{Appendix~\ref{sec:output-parsing}}.

\subsection{Latent Space Analysis}
\label{sec:latent-space}

We extract a hidden representation for every example at every model layer. For the primary layer-wise analysis, we apply KMeans to the normalized emotion-prompted final-position representations, setting \(k\) equal to the number of labels in the target taxonomy. We measure agreement between cluster assignments and GoEmotions labels using Adjusted Mutual Information (AMI), which measures similarity between two
partitions while correcting for agreement expected by chance. Higher AMI indicates closer correspondence between the discovered clusters and the target emotion labels, with 1 indicating identical partitions and values near 0 indicating approximately chance-level agreement. Because AMI depends on the structure of the target partition, we use it primarily for comparisons within the same taxonomy rather than to compare absolute scores across taxonomies.

To examine what level of organization the representations support without fixing \(k\) to the number of human-defined labels, we additionally sweep \(k=2,\ldots,35\) at each model's strongest 28-way layer. We measure cluster separation using silhouette, stability across random initializations using Adjusted Rand Index (ARI), and agreement with the target labels using AMI. We also repeat the 28-way sweep at the final layer as a robustness check. We record how many clusters are required for each taxonomy to reach near-maximal AMI.

Motivated by the cluster-count results, we conduct several follow-up analyses to characterize the resulting structure. We test whether the dominant two-cluster partition corresponds to broad sentiment or valence, examine whether fine-grained labels from the same emotion family co-cluster more often than labels from different families, and estimate the dimensionality of the emotion-centroid space. Additional definitions and controls are provided in
\hyperref[sec:latent-followup]{Appendix~\ref{sec:latent-followup}}.

Clustering measures whether label categories form distinct unsupervised groups, but weak clustering does not imply that label information is absent. We therefore train \(L_2\)-regularized linear classifiers on frozen hidden states and report held-out macro F1 for the 28-way, 7-way, and 4-way label spaces. The language-model parameters are never updated.

\xhdr{Representation Extraction}
\label{sec:representation-extraction}
Hidden representations can be extracted from an input in several ways. As an exploratory comparison using \texttt{Llama-3.2-1B-Instruct}, we compare raw-text and emotion-prompted inputs as well as several pooling strategies. Emotion prompting and final-position representations produce the strongest alignment with the fine-grained labels among the configurations tested. We therefore use emotion-prompted final-position representations for the cross-model analyses. Full comparisons are reported in \hyperref[sec:extraction-details]{Appendix~\ref{sec:extraction-details}}.


\section{Results}
We first ask how performance changes with label granularity and whether the hidden-state geometry supports a similarly coarse organization
(\hyperref[sec:coarse-structure]{Section~\ref{sec:coarse-structure}}). We then examine which fine-grained emotion distinctions overlap and whether those distinctions remain linearly recoverable (\hyperref[sec:fine-grained]{Section~\ref{sec:fine-grained}}).

\begin{table*}[t]
\centering
\begin{tabular}{lcccc}
\toprule
\textbf{Model}
& \textbf{28-way, no list}
& \textbf{28-way, list}
& \textbf{7-way}
& \textbf{4-way sentiment} \\
\midrule
Llama-3.2-1B & 0.089 {\footnotesize(49.3)} & 0.055 {\footnotesize(19.4)} & 0.113 {\footnotesize(5.1)} & 0.219 {\footnotesize(0.0)} \\
Llama-3.2-3B & 0.127 {\footnotesize(20.7)} & 0.225 {\footnotesize(1.9)}  & 0.338 {\footnotesize(1.3)} & 0.357 {\footnotesize(0.0)} \\
Llama-3.1-8B & 0.201 {\footnotesize(28.6)} & 0.242 {\footnotesize(1.6)}  & 0.335 {\footnotesize(0.2)} & 0.351 {\footnotesize(0.0)} \\
Gemma-2-2B   & 0.136 {\footnotesize(22.7)} & 0.208 {\footnotesize(1.0)}  & 0.329 {\footnotesize(2.0)} & 0.410 {\footnotesize(0.5)} \\
Gemma-2-9B   & 0.155 {\footnotesize(36.5)} & 0.250 {\footnotesize(1.5)}  & 0.347 {\footnotesize(0.8)} & 0.413 {\footnotesize(0.1)} \\
Qwen2.5-3B   & 0.167 {\footnotesize(18.9)} & 0.245 {\footnotesize(17.5)} & 0.349 {\footnotesize(3.1)} & 0.433 {\footnotesize(0.0)} \\
Qwen2.5-7B   & 0.168 {\footnotesize(18.5)} & 0.289 {\footnotesize(2.9)}  & 0.366 {\footnotesize(1.8)} & 0.409 {\footnotesize(0.0)} \\
\bottomrule
\end{tabular}
\caption{
\textbf{Output performance across models and requested label granularities.} Values show macro F1, with the percentage of unmatched outputs in parentheses. Unknown outputs are retained as errors when computing macro F1. Coarser requested label spaces generally yield higher F1 and fewer unmatched outputs,
although the mapped-output control in Appendix~\ref{sec:mapped-output} shows that this improvement does not by itself establish a coarser internal representation.
}
\label{tab:output-accuracy}
\end{table*}


\subsection{Coarse Structure Dominates Outputs and Latent Geometry}
\label{sec:coarse-structure}

Fine-grained emotion prediction is difficult across all seven models as shown in \hyperref[tab:output-accuracy]{Table~\ref{tab:output-accuracy}}. Without a candidate list, 28-way accuracy ranges from 8.2\% to 16.1\%, with macro F1 from .089 to .201. Providing the complete 28-label vocabulary raises accuracy in the six larger models, reaching 20.6-26.5\%, but fine-grained macro F1 remains between .225 and .289 in those models. Unmatched responses are also common in the unconstrained condition, accounting for approximately 19-49\% of outputs.

Directly requesting coarser labels substantially increases output performance. Among the six larger models, seven-way accuracy reaches 35.9-43.9\% and
four-way sentiment accuracy reaches 45.9-49.6\%. Four-way macro F1 ranges from .351 to .433 in these models, and almost all responses can be mapped to one of the requested sentiment labels. The smaller \texttt{Llama-3.2-1B-Instruct} remains an outlier, with four-way macro F1 of .219.

However, this improvement does not by itself show that the models represent emotion more strongly at coarse granularity. When predictions from the 28-way-list condition are instead mapped after generation into the coarser taxonomies, much of the apparent advantage remains (\hyperref[sec:mapped-output]{Appendix~\ref{sec:mapped-output}}). Among the six larger models, mapped 28-way predictions match or exceed directly prompted four-way macro F1 in five of six cases. Thus, the output results reflect both emotion recognition and the difficulty of producing the requested label vocabulary; we therefore rely on the hidden-state analyses to assess representational granularity more directly.

We first examine alignment between KMeans clusters and the 28 GoEmotions labels across model depth. Peak 28-way AMI ranges from .171 to .213 and occurs late in every model, at relative depth .76-.94. There is no consistent trend with parameter count: peak AMI increases across the three Llama models, falls from Gemma-2-2B to Gemma-2-9B, and is nearly unchanged across the two Qwen models.

Supplying the complete label list affects outputs much more strongly than it affects hidden-state organization. In the six larger models, output accuracy
increases substantially when the list is supplied, whereas peak 28-way latent AMI changes only slightly. This suggests that providing the vocabulary mainly
helps the models produce an allowed emotion name rather than substantially reorganizing the representation of the utterance.

\begin{figure*}[t]
    \centering
    \includegraphics[width=\textwidth]{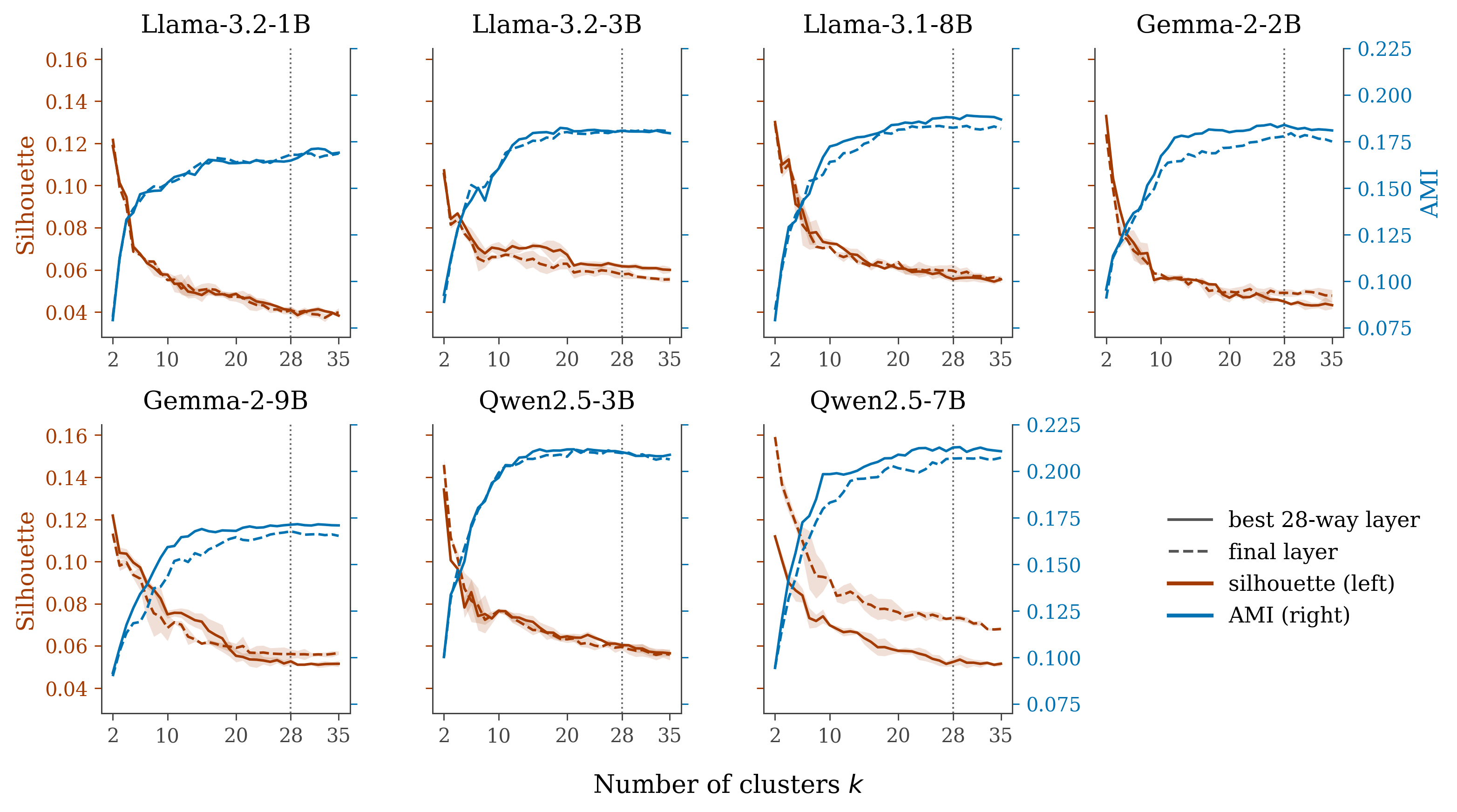}
    \caption{\textbf{The strongest unsupervised partition is substantially
    coarser than the 28-category GoEmotions taxonomy.} Each panel varies the KMeans cluster count at the model's strongest 28-way layer (solid) and final layer (dashed). Silhouette is highest at \(k=2\) in all seven models, whereas Adjusted Mutual Information (AMI) with the fixed 28-way labels rises as additional clusters are introduced. Taxonomy-specific \(k_{95}\) values are reported in Appendix~\ref{sec:latent-followup}.}
    \label{fig:k-sweep-crossmodel}
\end{figure*}

Fixing \(k\) to the number of GoEmotions labels does not reveal what granularity the representations support most strongly on their own. We therefore vary \(k=2,\ldots,35\) (\hyperref[fig:k-sweep-crossmodel]{Figure~\ref{fig:k-sweep-crossmodel}}).
Silhouette is highest at \(k=2\) in every model. Follow-up analysis shows that these two-cluster partitions are consistently related to valence: after matching the arbitrary cluster identities to positive and negative examples, balanced accuracy ranges from .783 to .837 across models. Neutral and ambiguous examples do not follow a purely binary split, so we describe this
partition as \emph{valence-related} rather than as an exact sentiment partition.

The same analysis shows a systematic relationship between taxonomy granularity and the number of clusters needed to recover its structure. The 28-way taxonomy requires \(k=13\)-17 to reach 95\% of its peak AMI, whereas the seven-way grouping requires only \(k=3\)-6 and the four-way sentiment grouping only \(k=2\)-4. These comparisons are made within each taxonomy and
do not rely on comparing absolute AMI values across taxonomies. Together, they show that broader affective distinctions can be recovered with substantially
fewer clusters than the full fine-grained label set.

This coarse organization does not imply that the emotion representation is itself one- or two-dimensional. At the layer of strongest 28-way alignment,
the mean-centered space spanned by the 28 emotion centroids has an effective rank of roughly 8-12 dimensions across models (\hyperref[sec:latent-followup]{Appendix~\ref{sec:latent-followup}}). The dominant coarse partition therefore
coexists with a substantially richer multidimensional structure.

\subsection{Fine-Grained Distinctions Co-Cluster Systematically}
\label{sec:fine-grained}

Overall alignment with the GoEmotions labels does not reveal which fine-grained distinctions account for the mismatch. We therefore ask whether emotions belonging to the same broader family are more likely to occupy the same clusters than emotions from different families.


The same three-level pattern appears in all seven models. At each model's strongest 28-way layer with \(k=28\), examples sharing the same fine-grained
label co-cluster with mean probability .140-.176. Examples with different fine-grained labels from the same broader family co-cluster with probability
.050-.060, compared with .027-.031 for labels from different families. Thus, same-family label pairs co-cluster \(1.7\)-\(2.2\times\) as often as cross-family pairs. The difference is significant in every model under 10,000 family-size-preserving permutations (\(p=.0001\)).

Because the emotion families contain different numbers of fine-grained labels, we also repeat the analysis giving each family equal weight rather than each
label pair equal weight. The within-family advantage remains in every model and becomes larger under this weighting, showing that the pattern is not driven by the larger joy family. Together, these results show that disagreement with the 28-way taxonomy is structured: labels belonging to the same broader
emotion family tend to overlap more strongly in the unsupervised geometry.

\begin{figure*}[t]
    \centering
    \includegraphics[width=\textwidth]{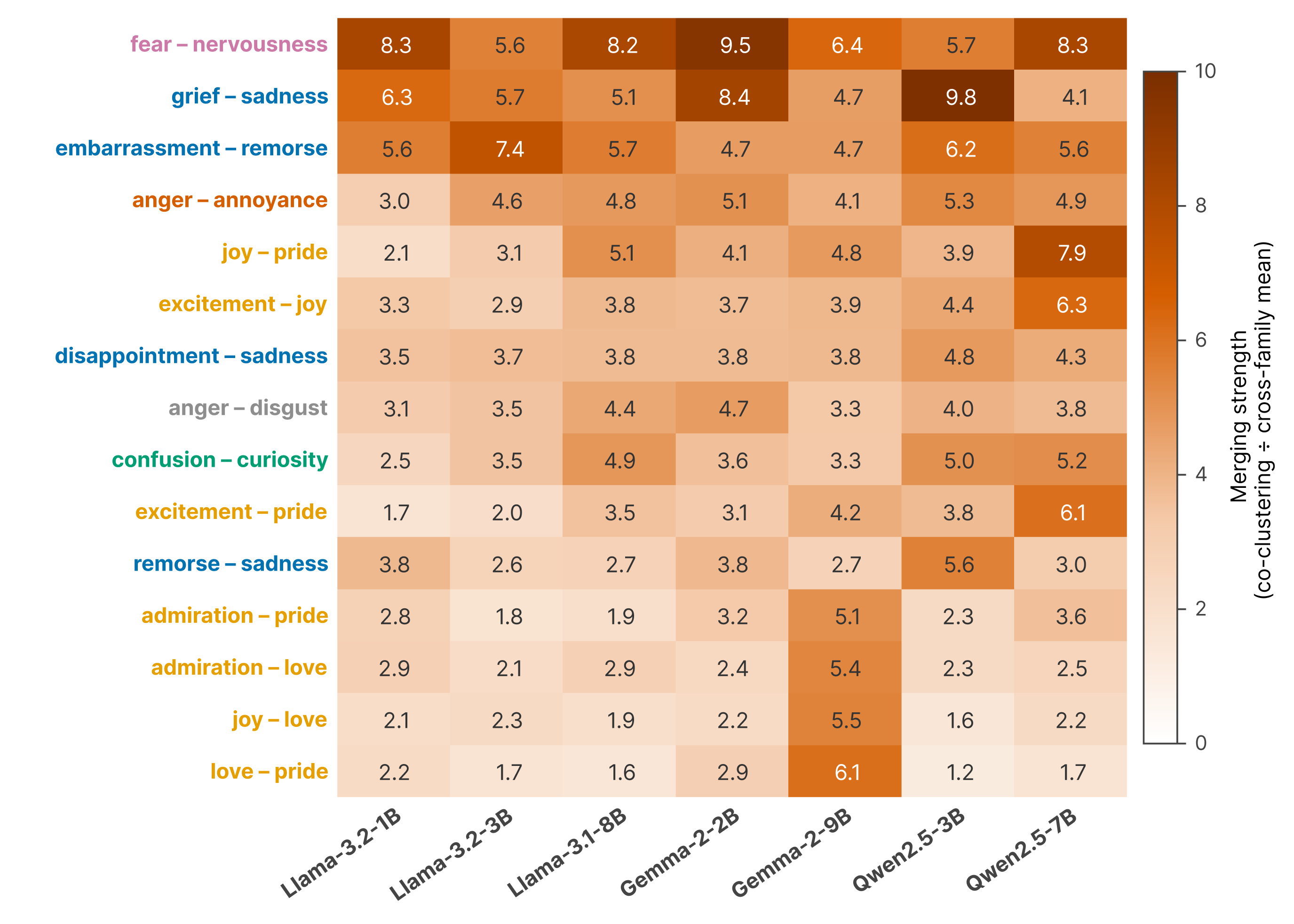}
    \caption{\textbf{Several fine-grained emotion pairs co-cluster strongly across models.} Rows include every label pair that ranks among the six most strongly co-clustered of the 378 different-label pairs in at least one model. Cells show normalized co-clustering strength: the pair's co-clustering probability divided by that model's mean co-clustering probability for cross-family pairs (e.g., 8.3 means the pair co-clusters 8.3 times as often as the average).}
    \label{fig:merged-pairs-crossmodel}
\end{figure*}

Several specific pairs recur across architectures
(\hyperref[fig:merged-pairs-crossmodel]{Figure~\ref{fig:merged-pairs-crossmodel}}). Using the same co-clustering scores to rank all 378 different-label pairs within each model, fear--nervousness ranks among the three strongest pairs in every model, embarrassment--remorse among the top seven, and grief--sadness among the top eleven. The cross-family pair anger--disgust also consistently shows above-baseline co-clustering, indicating that the learned organization does not exactly reproduce the seven-way family boundaries. Some larger models also show strong co-clustering among positive pairs such as joy--pride, excitement--joy, love--pride, and joy--love. Because several of these labels are rare in the dataset, individual pairwise results should be interpreted more cautiously than the aggregate family-level effect.

The same family-level pattern also appears without using cluster assignments. Across all seven models, centroids for fine-grained labels from the same
family have higher cosine similarity than centroids from different families. Family-permutation tests are significant in every model (\(p\leq.001\) in six models; \(p=.0115\) for \texttt{Gemma-2-2B-IT}). This provides a clustering-independent indication that labels within the same broader family tend to occupy more similar regions of representation space.

\begin{table}[H]
\centering
\begin{tabular}{lccc}
\toprule
\textbf{Model} & \textbf{28-way} & \textbf{7-way} & \textbf{4-way} \\
\midrule
Llama-3.2-1B & 0.376 & 0.540 & 0.632 \\
Llama-3.2-3B & 0.404 & 0.556 & 0.638 \\
Llama-3.1-8B & 0.378 & 0.524 & 0.642 \\
Gemma-2-2B   & 0.368 & 0.520 & 0.629 \\
Gemma-2-9B   & 0.360 & 0.524 & 0.625 \\
Qwen2.5-3B   & 0.365 & 0.514 & 0.623 \\
Qwen2.5-7B   & 0.401 & 0.547 & 0.654 \\
\bottomrule
\end{tabular}
\caption{
\textbf{Fine-grained label information remains linearly recoverable from frozen model representations.} Held-out macro F1 of regularized linear probes at each model's strongest
28-way clustering layer. Although 28-way labels do not form clean unsupervised clusters, they remain substantially predictable with a linear readout. Performance increases for the seven-way and four-way label spaces.
}
\label{tab:probe-results}
\end{table}

The clustering results do not imply that fine-grained label information is absent. Linear classifiers trained on frozen hidden states recover 28-way labels with macro F1 between 0.360 and 0.404 (\hyperref[tab:probe-results]{Table~\ref{tab:probe-results}}). Performance rises to .514-.556 for the
seven-way taxonomy and .623-.654 for four-way sentiment. Fine-grained label information therefore remains linearly recoverable even though the categories
do not form cleanly separated regions under the unsupervised clustering analysis.

Finally, the family-level co-clustering effect is stable across nearby layers, cluster counts, and KMeans initializations in all seven models. It is weaker
and less consistent under average-linkage agglomerative clustering, indicating that the magnitude and ranking of individual pairwise effects depend partly on
the clustering method. Full robustness results are reported in \hyperref[sec:clustering-robustness]{Appendix~\ref{sec:clustering-robustness}}.

\section{Discussion}

Across all seven models, the dominant unsupervised organization of emotion-prompted representations is substantially coarser than the 28-category
GoEmotions taxonomy. We use \emph{overcompression} to describe this pattern: broad affective distinctions, especially valence-related structure, organize
the representations more strongly than many of the finer distinctions encoded by the dataset. The systematic co-clustering of labels within the same emotion families further shows that this mismatch is structured rather than arbitrary.

There are at least two ways to interpret this result. One possibility is that models represent some fine-grained emotion distinctions only weakly. Another
is that those distinctions are present but embedded in a continuous or hierarchical space that is not well captured by discrete KMeans clusters. Our linear-probe results support the second interpretation in part: substantial 28-way label information remains linearly recoverable even when the same labels do not form clean unsupervised clusters. Thus, supervised recoverability and unsupervised geometric separation should not be treated as equivalent measures of whether a model represents an emotion distinction.

This distinction also helps connect our findings with prior work showing that LLMs contain coherent or causally relevant emotion concepts
\citep{tak2025mechanistic,lee2025emotionneurons,sofroniew2026emotion,reichman2026emotions}.
A model may contain information associated with concepts such as fear, nervousness, or grief without organizing naturally occurring utterances into
sharply separated regions corresponding to those labels. We therefore distinguish \emph{concept availability} from \emph{utterance organization}.
This interpretation is also compatible with hierarchical accounts of emotion structure in model outputs \citep{okawa2026hierarchical}.

The pattern is not explained by model size within the range we study: fine-grained alignment does not improve monotonically from approximately 1B to 9B parameters. This is a descriptive comparison rather than a controlled scaling result, however. The measured geometry also depends on the representation protocol and clustering method. Emotion-focused prompting and readout position substantially affect measured alignment, and the family-level effect is less consistent under agglomerative clustering than under KMeans. Claims that a model ``represents'' an emotion should therefore specify whether they refer to decodable information, unsupervised geometric organization, or causally used information.

Our experiments do not determine which of these forms of representation is most relevant to model behavior. Distinguishing weak fine-grained representation from a richer continuous or hierarchical organization will
require comparisons with human similarity judgments, broader geometric analyses, and causal tests of whether particular distinctions affect generation. Likewise, although distinctions such as fear versus nervousness or
grief versus sadness could matter in downstream interactions, we do not test response quality or safety consequences here.

\section{Conclusion}

We studied how seven open instruction-tuned language models organize emotions expressed in 36,308 GoEmotions examples. Under an emotion-prompted hidden-state protocol, the strongest unsupervised structure is consistently coarser than the 28-category GoEmotions taxonomy: the dominant two-cluster partition is valence-related, coarser taxonomies require fewer clusters to
recover most of their available alignment, and related fine-grained labels co-cluster systematically. At the same time, linear probes retain substantial predictive information about the 28 fine-grained labels.

We describe this pattern as \emph{overcompression} relative to the GoEmotions
taxonomy. Fine-grained label information can remain recoverable even when it
does not form cleanly separated regions in the dominant unsupervised geometry.

\section*{Limitations}

We evaluate seven open instruction-tuned models spanning three families and approximately 1-9B parameters. This range does not include substantially
larger open models or closed frontier systems, for which intermediate hidden states are generally unavailable. We observe no consistent relationship
between parameter count and fine-grained alignment within the models studied, but this should not be interpreted as a controlled scaling result.

Our reference structure is the GoEmotions label taxonomy rather than a direct measurement of human emotion geometry. Because we retain only examples with a
single released label, we cannot compare model co-clustering directly with how individual annotators relate or confuse pairs of emotions. This is especially
relevant for pairs such as fear-nervousness: they co-cluster strongly across models, but \textit{nervousness} also has relatively low inter-annotator agreement in GoEmotions \citep{demszky2020goemotions}. Our results therefore
establish a mismatch with the dataset taxonomy, not with human psychological representation itself.

The single-label restriction also excludes mixed-affect examples, where several emotion labels are assigned to the same utterance. These examples may be especially informative about boundaries between fine-grained emotions, so our results do not characterize that part of the dataset. In addition, several fine-grained labels are rare, making individual pairwise co-clustering ranks less reliable than the aggregate family-level patterns.

Our conclusions also depend on methodological choices. Emotion prompting, readout position, and pooling strategy substantially change measured label alignment, and the family-level co-clustering effect is less consistent under average-linkage agglomerative clustering than under KMeans. The \(k=2\) partitions are consistently valence-related, but they do not form an exact positive-versus-negative split, particularly for neutral and ambiguous examples. We therefore interpret them as evidence of a dominant coarse affective partition rather than as exactly two emotion categories.

Class imbalance may also influence the unsupervised partitions because KMeans is fit to the naturally imbalanced dataset; we do not repeat the full cross-model analysis on class-balanced or neutral-excluded subsets. In addition, the agglomerative robustness check changes both the clustering algorithm and the distance metric, so it does not isolate which methodological choice accounts for the difference.

Finally, clustering and probing answer different questions. Clustering tests whether labels form distinct unsupervised regions, whereas linear probes test whether label information can be recovered by a learned readout. Neither establishes that the model causally uses these distinctions during generation. Our probes are also global multiclass probes rather than within-family probes, so they do not directly establish whether close distinctions such as
fear-nervousness remain linearly separable. Moreover, because probe results are reported at layers selected for strong 28-way clustering alignment, they should be interpreted as descriptive rather than as a fully nested estimate of layer selection on unseen data. Intervention-based experiments and leak-free layer-selection procedures would provide stronger evidence about causal use and generalization.

\section*{Ethical Considerations}
Our analysis concerns representations of emotion expressed in text, not emotions
exhibited by the models themselves. A risk of this line of work is
overinterpreting geometric or probe-based evidence as evidence of human-like
emotional understanding. Our reference structure is the GoEmotions taxonomy,
which reflects a particular annotation scheme and dataset rather than a
universal account of human emotion. Care should therefore be taken not to treat
differences between model representations and this taxonomy as lack of psychological knowledge, or to generalize the observed structure beyond what is represented by the dataset.

Methods for identifying or characterizing emotion representations could also
be used in systems that infer affective states from user text. Such systems may
enable profiling, surveillance, targeted persuasion, or automated decisions in
sensitive settings such as healthcare, education, employment, or moderation.
These risks are especially important because emotion expression varies across
individuals and cultural contexts, while our experiments use English Reddit
comments and do not evaluate demographic or cross-cultural variation. We
therefore do not recommend using the distinctions studied here as a basis for judgments about individuals.
\\\\
Disclosure: The authors used LLMs to assist with revising portions of the manuscript and LaTeX formatting across the paper including  tables. All results and claims were reviewed and verified by the authors.


\bibliography{custom}

\appendix


\section{Representation Extraction Details}
\label{sec:extraction-details}

We use \texttt{Llama-3.2-1B-Instruct} for an exploratory comparison of
representation-extraction choices before fixing a common protocol for the
cross-model experiments. We vary both the input context and the method used
to reduce token-level hidden states to one representation per example.

\paragraph{Tokenization and readout.}
For emotion-prompted representations, we render each prompt using the model's
native chat template with \texttt{add\_generation\_prompt=True}. We tokenize
the complete rendered sequence and truncate inputs to a maximum length of 384
tokens. At every layer, we extract the hidden state at the final non-padding
input position and \(L_2\)-normalize the resulting vector before clustering.
We refer to this as the \emph{final-position representation}. Because the
native chat template may append model-specific special tokens, this position
should not be interpreted as necessarily corresponding to the literal final
character of the written prompt.

\paragraph{Prompt conditioning.}
We first compare representations extracted from the original utterance alone
with representations extracted after prompting the model to identify the
emotion expressed in the utterance. Emotion-focused prompting increases
best-layer AMI under all four target spaces considered
(\hyperref[tab:prompt-comparison]{Table~\ref{tab:prompt-comparison}}). The increase is largest in the 28-way
no-list condition, where AMI rises from .106 to .171.

\begin{table}[H]
\centering
\small
\begin{tabular}{lcc}
\toprule
\textbf{Condition} & \textbf{Text only} & \textbf{Emotion prompt} \\
\midrule
28-way, no list     & .106 & .171 \\
28-way, list given  & .106 & .135 \\
7-way               & .070 & .104 \\
4-way sentiment     & .056 & .129 \\
\bottomrule
\end{tabular}
\caption{\textbf{Effect of emotion-focused prompting on latent-space
alignment.} Best-layer AMI for \texttt{Llama-3.2-1B-Instruct} using
raw-text input or the emotion-identification prompt.}
\label{tab:prompt-comparison}
\end{table}

\paragraph{Pooling strategy.}
We next compare the final-position hidden state with mean pooling over the
utterance tokens and mean pooling over the complete prompted sequence
(\hyperref[tab:pooling-comparison]{Table~\ref{tab:pooling-comparison}}). Final-position representations preserve
substantially more GoEmotions-label alignment than either mean-pooling
strategy. In the 28-way no-list condition, best-layer AMI falls from .171
with the final-position representation to .045 with text-mean pooling and
.042 with full-prompt mean pooling.
\\\\
These comparisons motivate the emotion-prompted final-position
representation used in the cross-model analyses. Because this comparison is
performed only on \texttt{Llama-3.2-1B-Instruct}, we treat the selected
protocol as a methodological choice rather than evidence that it is optimal
for every model.

\begin{table}[H]
\centering
\small
\begin{tabular}{lccc}
\toprule
\textbf{Condition}
& \textbf{Final Pos}
& \textbf{Text mean}
& \textbf{FP mean} \\
\midrule
28-way, no list   & .171 & .045 & .042 \\
4-way sentiment   & .129 & .010 & .003 \\
\bottomrule
\end{tabular}
\caption{\textbf{Effect of pooling strategy on latent-space alignment.}
Best-layer AMI for \texttt{Llama-3.2-1B-Instruct} under the emotion-focused
prompt. FP mean stands for Full-prompt mean.}
\label{tab:pooling-comparison}
\end{table}


\section{Prompts}
\label{sec:prompts}

\paragraph{28-way, no list.}
\begin{quote}
\small
\texttt{Identify the single emotion expressed in this text. Respond with exactly one emotion word. Do not explain.}\\
\texttt{Text: [TEXT]}\\
\texttt{Emotion:}
\end{quote}

\paragraph{28-way, list given.}
\begin{quote}
\small
\texttt{Choose exactly one label from this list: [28 GOEMOTIONS LABELS].}\\
\texttt{Output only the label name, nothing else.}\\
\texttt{Text: [TEXT]}\\
\texttt{Label:}
\end{quote}

\paragraph{7-way.}
\begin{quote}
\small
\texttt{Choose exactly one emotion from: anger, disgust, fear, joy, sadness, surprise, neutral.}\\
\texttt{Output only the emotion name, nothing else.}\\
\texttt{Text: [TEXT]}\\
\texttt{Emotion:}
\end{quote}

\paragraph{4-way sentiment.}
\begin{quote}
\small
\texttt{Choose exactly one sentiment from: positive, negative, ambiguous, neutral.}\\
\texttt{Output only the sentiment, nothing else.}\\
\texttt{Text: [TEXT]}\\
\texttt{Sentiment:}
\end{quote}

\section{Output Generation and Parsing}
\label{sec:output-parsing}

\paragraph{Generation.}
We generate deterministically with \texttt{do\_sample=False} and
\texttt{max\_new\_tokens=25}, using each model's native tokenizer and chat
template.

\begin{table*}[t]
\centering
\small
\begin{tabular}{lcccc}
\toprule
\textbf{Model}
& \textbf{28-way, no list}
& \textbf{28-way, list}
& \textbf{7-way}
& \textbf{4-way sentiment} \\
\midrule
Llama-3.2-1B & .089 & .055 & .113 & .219 \\
Llama-3.2-3B & .127 & .225 & .338 & .357 \\
Llama-3.1-8B & .201 & .242 & .335 & .351 \\
Gemma-2-2B   & .136 & .208 & .329 & .410 \\
Gemma-2-9B   & .155 & .250 & .347 & .413 \\
Qwen2.5-3B   & .167 & .245 & .349 & .433 \\
Qwen2.5-7B   & .168 & .289 & .366 & .409 \\
\bottomrule
\end{tabular}
\caption{\textbf{Corrected macro F1 across output-space conditions.}
Unmatched responses remain in the evaluation and count as errors rather than
being filtered before computing F1.}
\label{tab:output-f1}
\end{table*}

\begin{table*}[t]
\centering
\small
\begin{tabular}{lcccc}
\toprule
\textbf{Model}
& \textbf{28-way, no list}
& \textbf{28-way, list}
& \textbf{7-way}
& \textbf{4-way sentiment} \\
\midrule
Llama-3.2-1B & 49.3 & 19.4 & 5.1 & 0.0 \\
Llama-3.2-3B & 20.7 & 1.9 & 1.3 & 0.0 \\
Llama-3.1-8B & 28.6 & 1.6 & 0.2 & 0.0 \\
Gemma-2-2B   & 22.7 & 1.0 & 2.0 & 0.5 \\
Gemma-2-9B   & 36.5 & 1.5 & 0.8 & 0.1 \\
Qwen2.5-3B   & 18.9 & 17.5 & 3.1 & 0.0 \\
Qwen2.5-7B   & 18.5 & 2.9 & 1.8 & 0.0 \\
\bottomrule
\end{tabular}
\caption{\textbf{Unknown-response rate (\%) across output-space conditions.}
Unmatched responses are most common when models must freely produce a
fine-grained emotion name and fall sharply when a candidate vocabulary is
provided.}
\label{tab:output-unknown}
\end{table*}

\paragraph{Parsing.}
Generated outputs are lowercased and stripped of leading and trailing
non-word characters. We first attempt an exact match against the permitted
labels, followed by an exact match on the first generated word. For the
unconstrained 28-way condition, we additionally apply a predefined
emotion-synonym mapping to the complete response and to individual generated
words. We then search for permitted labels as substrings, checking longer
labels first. Responses that cannot be mapped by any of these rules are marked
as \emph{unknown}.

Unknown predictions are retained as errors when computing both total accuracy
and macro F1; they are not removed before scoring. We additionally report the
unknown-response rate because the frequency of unmappable outputs differs
substantially across prompting conditions.

Manual inspection of unmatched outputs indicates that an \emph{unknown}
response is not equivalent to a pure emotion-recognition error. Unmatched
generations include plausible emotion terms outside the requested taxonomy
(e.g., near-synonyms or related affective terms) as well as outputs that do not
follow the requested response format. We therefore interpret unknown rate as
reflecting both emotion prediction and vocabulary/instruction alignment.


\section{Additional Output-Space Results}
Corrected macro F1 scores for all output conditions are reported in
\hyperref[tab:output-f1]{Table~\ref{tab:output-f1}}. The corresponding unknown-response rates are reported in \hyperref[tab:output-unknown]{Table~\ref{tab:output-unknown}}.

\label{sec:output-additional}

\subsection{Mapped Output-Space Control}
\label{sec:mapped-output}

Directly prompting a model with a coarser label space changes both the target
taxonomy and the generation task. To separate these effects, we take
predictions from the 28-way-list condition and map the predicted fine-grained
label to its corresponding seven-way and four-way category after generation.
We then compare these mapped predictions with predictions produced when the
model is directly prompted for the coarse label
(\hyperref[tab:mapped-output]{Table~\ref{tab:mapped-output}}).

\begin{table*}[t]
\centering
\small
\begin{tabular}{lcccc}
\toprule
\textbf{Model}
& \textbf{Direct 7-way}
& \(\mathbf{28\rightarrow7}\)
& \textbf{Direct 4-way}
& \(\mathbf{28\rightarrow4}\) \\
\midrule
Llama-3.2-3B & .338 & .334 & .357 & .376 \\
Llama-3.1-8B & .335 & .375 & .351 & .438 \\
Gemma-2-2B   & .329 & .371 & .410 & .488 \\
Gemma-2-9B   & .347 & .370 & .413 & .440 \\
Qwen2.5-3B   & .349 & .357 & .433 & .398 \\
Qwen2.5-7B   & .366 & .360 & .409 & .433 \\
\bottomrule
\end{tabular}
\caption{\textbf{Macro F1 when coarse labels are requested directly or
obtained by mapping a 28-way prediction after generation.}
For five of the six larger models, mapping the 28-way prediction to four-way
sentiment matches or exceeds the directly prompted four-way result. Output
improvements under coarse prompting therefore cannot by themselves establish
that the internal representation is coarser.}
\label{tab:mapped-output}
\end{table*}


\section{Additional Latent-Space Analyses}
\label{sec:latent-followup}

\subsection{Cluster-Count Summary}

For each target taxonomy, we define \(k_{95}\) as the smallest number of
clusters whose AMI reaches at least 95\% of the maximum AMI observed over
\(k=2,\ldots,35\). Because AMI depends on the target partition, \(k_{95}\)
is computed separately within each taxonomy; we do not interpret differences
in absolute AMI across the 28-way, seven-way, and four-way targets.

\subsection{Interpreting the Two-Cluster Partition}

To characterize the dominant \(k=2\) partition, we compare its assignments
with the GoEmotions sentiment grouping. We measure association with the full
four-way positive, negative, ambiguous, and neutral labels using Cramér's
\(V\). We then restrict the analysis to positive and negative examples and
compare the two arbitrary cluster IDs with binary valence after choosing the
cluster-to-valence assignment that maximizes agreement. We report balanced accuracy and AMI for this binary comparison
(\hyperref[tab:k2-valence]{Table~\ref{tab:k2-valence}}).

We repeat KMeans under three random initializations and measure agreement
between the resulting clusterings with ARI. The two-cluster partition is
consistently valence-related across all seven models, although neutral and
ambiguous examples prevent interpreting it as an exact binary sentiment
partition.

\begin{table*}[t]
\centering
\small
\begin{tabular}{lcccc}
\toprule
\textbf{Model}
& \textbf{Valence AMI}
& \textbf{Balanced Acc.}
& \textbf{4-way Cramér's \(V\)}
& \textbf{Cross-seed ARI} \\
\midrule
Llama-3.2-1B & .263 & .798 & .430 & .779 \\
Llama-3.2-3B & .248 & .783 & .449 & .685 \\
Llama-3.1-8B & .263 & .799 & .449 & .837 \\
Gemma-2-2B   & .313 & .812 & .496 & .992 \\
Gemma-2-9B   & .289 & .802 & .479 & .926 \\
Qwen2.5-3B   & .352 & .837 & .509 & .948 \\
Qwen2.5-7B   & .316 & .820 & .496 & .845 \\
\bottomrule
\end{tabular}
\caption{\textbf{The dominant two-cluster partition is consistently
valence-related.} Binary-valence metrics use positive and negative examples;
Cramér's \(V\) uses the full four-way sentiment grouping.}
\label{tab:k2-valence}
\end{table*}

\subsection{Fine-Grained Co-Clustering}

At each model's strongest 28-way layer with \(k=28\), we compute the
probability that examples belonging to each pair of fine-grained labels are
assigned to the same cluster. We distinguish three cases: examples sharing
the same fine-grained label, examples with different labels belonging to the
same seven-way family, and examples whose labels belong to different
families.

To test whether the same-family advantage exceeds chance under the observed
family structure, we use a one-sided permutation test. Fine-grained labels
are randomly reassigned to families while preserving the original family-size
multiset. The test statistic is the difference between mean within-family and
mean cross-family co-clustering probability. We use 10,000 permutations and a
plus-one Monte Carlo correction.

The seven-way families contain different numbers of fine-grained labels, so a
pair-weighted mean gives greater influence to large families, especially joy.
We therefore additionally compute an equal-family-weighted summary in which
each multi-label family contributes equally
(\hyperref[tab:family-weighting]{Table~\ref{tab:family-weighting}}).

\begin{table}[H]
\centering
\small
\begin{tabular}{lcc}
\toprule
\textbf{Model}
& \textbf{Pair-weighted gap}
& \textbf{Equal-family gap} \\
\midrule
Llama-3.2-1B & .025 & .077 \\
Llama-3.2-3B & .020 & .060 \\
Llama-3.1-8B & .026 & .078 \\
Gemma-2-2B   & .026 & .092 \\
Gemma-2-9B   & .030 & .062 \\
Qwen2.5-3B   & .029 & .072 \\
Qwen2.5-7B   & .032 & .077 \\
\bottomrule
\end{tabular}
\caption{\textbf{Same-family minus cross-family co-clustering gaps under two
weighting schemes.} Giving each family equal weight strengthens the effect in
every model, indicating that the aggregate result is not driven by the larger
joy family.}
\label{tab:family-weighting}
\end{table}

\subsection{Emotion-Centroid Dimensionality}

To characterize emotion-level dimensionality independently of KMeans
assignments, we compute one centroid for each of the 28 fine-grained labels by
averaging the representations of examples with that label. We subtract the
mean of the 28 centroids and apply singular value decomposition to the
resulting \(28\times d\) matrix. If the squared singular values are
\(\lambda_i\), we define

\[
p_i = \frac{\lambda_i}{\sum_j \lambda_j}.
\]

We summarize the resulting spectrum using the participation ratio,

\[
D_{\mathrm{PR}} = \frac{1}{\sum_i p_i^2},
\]

and entropy effective rank,

\[
D_{\mathrm{ER}}
=
\exp\left(-\sum_i p_i \log p_i\right).
\]

\begin{figure*}[t]
    \centering
    \includegraphics[width=\textwidth]{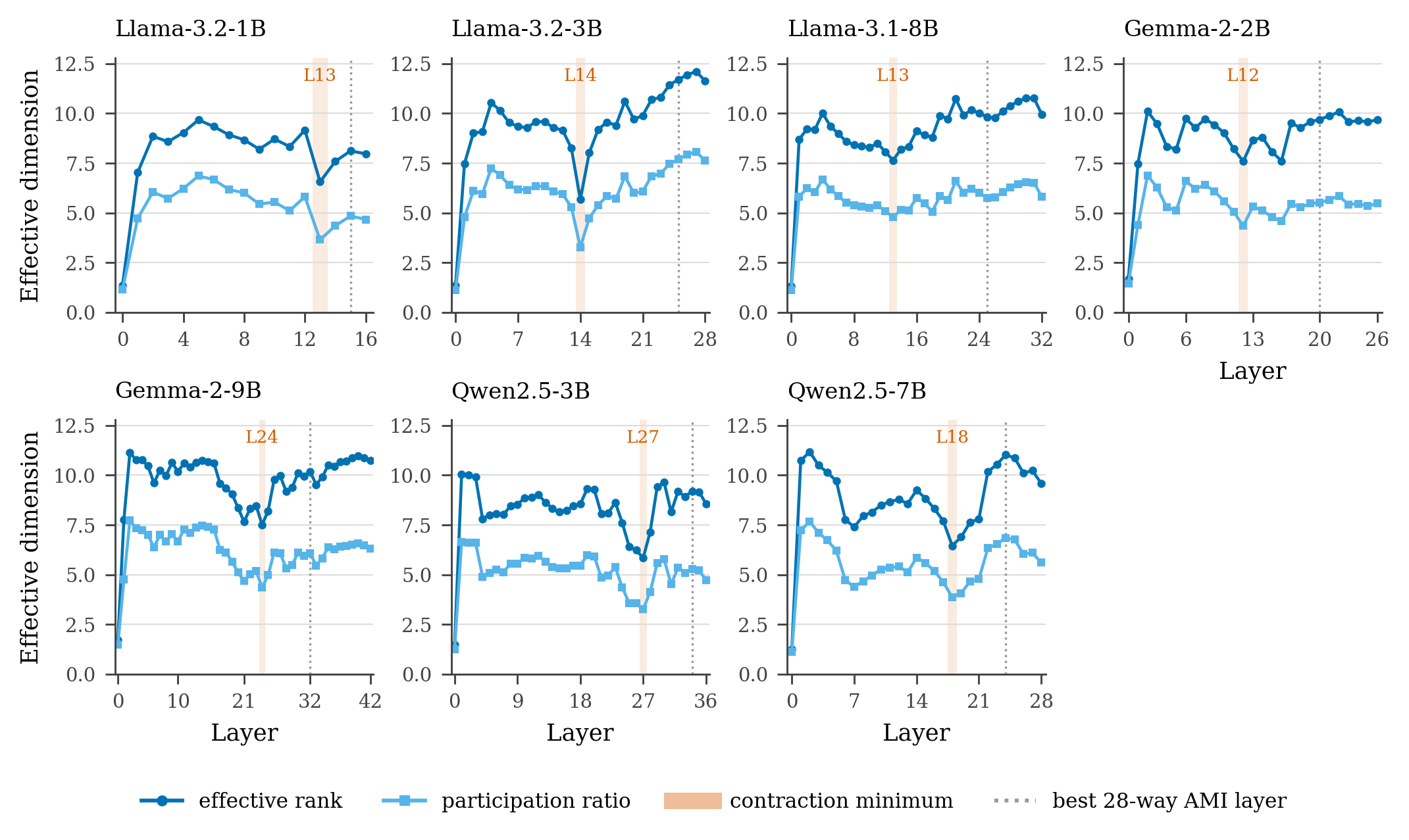}
    \caption{\textbf{Emotion-centroid dimensionality changes non-monotonically
    across depth.}
    Effective rank and participation ratio are computed from the 28
    mean-centered emotion centroids. The shaded band marks the minimum
    effective-rank layer, while the dotted line marks the layer of maximal
    28-way AMI. The centroid subspace remains multidimensional rather than
    collapsing to one or two dimensions.}
    \label{fig:dimensionality-crossmodel}
\end{figure*}

Because the centered centroid matrix has at most rank 27, these quantities
describe the dimensionality of the emotion-centroid subspace rather than the
full hidden-state space. Their trajectories across model depth are shown in
\hyperref[fig:dimensionality-crossmodel]{Figure~\ref{fig:dimensionality-crossmodel}}.

\section{Clustering Robustness}
\label{sec:clustering-robustness}

We test whether the family-level co-clustering effect depends on layer,
cluster count, KMeans initialization, or clustering procedure. For each model,
we evaluate the strongest 28-way layer and nearby layers, the final layer,
\(k\in\{7,14,20,28,35\}\), and KMeans seeds \(0,1,2\).

Across all seven models, the same-family minus cross-family co-clustering gap
is positive and statistically significant throughout the tested KMeans
configurations. Median correlations between pairwise co-clustering rankings
and the primary analysis range from .86 to .91, indicating that pair rankings
are also relatively stable under these KMeans perturbations.

We additionally repeat the analysis using average-linkage agglomerative
clustering with cosine distance on a fixed stratified subsample capped at 100
examples per fine-grained label. The family-level effect is less consistent:
the gap is positive in 35-93\% of configurations and statistically
significant in 10-80\%, depending on the model, with median pair-rank
correlations ranging from .03 to .69.

This robustness comparison changes both the clustering algorithm and the
distance metric relative to the primary Euclidean KMeans analysis. It
therefore demonstrates method dependence but does not isolate which of these
choices causes the difference. We treat the cross-model family effect as
robust within the primary KMeans protocol while avoiding claims that exact
pair rankings are invariant across clustering methods.


\section{Linear-Probe Details}
\label{sec:probe-details}

We train \(L_2\)-regularized multinomial linear classifiers on frozen hidden
representations; the language-model parameters are never updated. Examples are
partitioned using a fixed stratified 70/15/15 train/validation/test split based
on the original 28-way label. Regularization strength is selected using the
validation split, and final performance is reported as macro F1 on the held-out
test split for the 28-way, seven-way, and four-way targets. The resulting values for all seven models are reported in \hyperref[tab:k95]{Table~\ref{tab:k95}}.

\begin{table}[H]
\centering
\small
\begin{tabular}{lccc}
\toprule
\textbf{Model} & \(\mathbf{k_{95}}\) \textbf{28-way}
& \(\mathbf{k_{95}}\) \textbf{7-way}
& \(\mathbf{k_{95}}\) \textbf{4-way} \\
\midrule
Llama-3.2-1B & 17 & 3 & 2 \\
Llama-3.2-3B & 13 & 4 & 3 \\
Llama-3.1-8B & 15 & 4 & 4 \\
Gemma-2-2B   & 14 & 4 & 2 \\
Gemma-2-9B   & 15 & 6 & 2 \\
Qwen2.5-3B   & 13 & 3 & 3 \\
Qwen2.5-7B   & 15 & 6 & 3 \\
\bottomrule
\end{tabular}
\caption{\textbf{Number of clusters needed to reach 95\% of peak AMI within
each target taxonomy.} Coarser taxonomies reach near-maximal alignment using
substantially fewer clusters than the 28-way taxonomy.}
\label{tab:k95}
\end{table}

The main-text probe results are reported at each model's strongest 28-way
clustering layer. This layer is selected using 28-way clustering alignment
rather than probe test performance, but the layer-selection procedure is not
fully nested with respect to the held-out probe data. We therefore interpret
the reported probe results as evidence that fine-grained label information is
linearly recoverable, rather than as an unbiased estimate of the best
generalizing probe layer.

The global multiclass probes also do not directly test whether individual
within-family distinctions, such as fear versus nervousness, remain linearly
separable. Pairwise or within-family probes would provide a stronger test of
those local distinctions.

\section{Computational Resources}
The experiments involve inference and hidden-state extraction for seven
open-weight language models ranging from approximately 1B to 9B parameters
over 36,308 examples. We estimate that the primary experiments required
approximately 100 NVIDIA A100 GPU-hours in total. This estimate excludes
failed or exploratory runs and lightweight downstream analyses such as
clustering, permutation tests, and linear-probe fitting. Compute requirements
varied across models because of differences in model size, batch size, and
memory requirements. Computational resources and GPUs were provided by the authors' research institute.

\end{document}